\documentclass[letterpaper]{article} %
\usepackage{aaai26_arxiv}
\usepackage{times}  %
\usepackage{helvet}  %
\usepackage{courier}  %
\usepackage[hyphens]{url}  %
\usepackage{graphicx} %

\usepackage{subfiles} %
\usepackage{makecell} %
\usepackage{multirow} %
\usepackage{subcaption}
\usepackage{color,xcolor}
\usepackage{booktabs}
\usepackage{pifont}

\usepackage{amsmath}
\usepackage{amssymb}
\usepackage{mathtools}
\usepackage{amsthm}

\usepackage{natbib}  %
\usepackage{caption} %
\usepackage{array}
\newcolumntype{C}{>{\centering\arraybackslash}p{1.4cm}}
\usepackage{algorithm}
\usepackage{algorithmic}
\usepackage{adjustbox}

\usepackage{xcolor,colortbl}
\definecolor{citecolor}{HTML}{0071BC}
\definecolor{linkcolor}{HTML}{ED1C24}

\usepackage[pagebackref=false, breaklinks=true, colorlinks, citecolor=citecolor, linkcolor=linkcolor, bookmarks=false]{hyperref}

\usepackage{newfloat}
\usepackage{listings}
\DeclareCaptionStyle{ruled}{labelfont=normalfont,labelsep=colon,strut=off} %
\floatstyle{ruled}
\newfloat{listing}{tb}{lst}{}
\floatname{listing}{Listing}
\newcommand{\name}{LENS}

\title{\name: Learning to Segment Anything with Unified Reinforced Reasoning}
\author{
    Lianghui Zhu\textsuperscript{\rm 1}\equalcontrib,
    Bin Ouyang\textsuperscript{\rm 1}\equalcontrib\thanks{Work was done during Bin Ouyang's internship at vivo Mobile Communication Co., Ltd.},
    Yuxuan Zhang\textsuperscript{\rm 1},
    Tianheng Cheng\textsuperscript{\rm 1}\thanks{Project leader},
    Rui Hu\textsuperscript{\rm 1},
    Haocheng Shen\textsuperscript{\rm 2},
    Longjin Ran\textsuperscript{\rm 2},
    Xiaoxin Chen\textsuperscript{\rm 2},
    Li Yu\textsuperscript{\rm 1},
    Wenyu Liu\textsuperscript{\rm 1},
    Xinggang Wang\textsuperscript{\rm 1}\thanks{Corresponding author (\protect\url{xgwang@hust.edu.cn})}
}
\affiliations{
    \textsuperscript{\rm 1}School of EIC, Huazhong University of Science \& Technology,\\
    \textsuperscript{\rm 2}vivo Mobile Communication Co., Ltd
}

\begin{document}

\maketitle

\begin{abstract}
    Text-prompted image segmentation enables fine-grained visual understanding and is critical for applications such as human-computer interaction and robotics. 
    However, existing supervised fine-tuning methods typically ignore explicit chain-of-thought (CoT) reasoning at test time, which limits their ability to generalize to unseen prompts and domains.
    To address this issue, we introduce \name, a scalable reinforcement-%
    learning framework that jointly optimizes the reasoning process and segmentation in an end-to-end manner.
    We propose unified reinforcement-learning rewards that span sentence-, box-, and segment-level cues, encouraging the model to generate informative CoT rationales while refining mask quality.
    Using a publicly available 3-billion-parameter vision–language model, i.e., Qwen2.5-VL-3B-Instruct, \name{} achieves an average cIoU of 81.2\% on the RefCOCO, RefCOCO+, and RefCOCOg benchmarks, outperforming the strong fine-tuned method, i.e., GLaMM, by up to 5.6\%. %
    These results demonstrate that RL-driven CoT reasoning significantly enhances text-prompted segmentation and offers a practical path toward more generalizable Segment Anything models (SAM).

\end{abstract}

\begin{links}
    \link{Code \& Models}{https://github.com/hustvl/LENS}
    \link{Project Page}{https://hustvl.github.io/LENS/}
    \link{Extended version}{https://arxiv.org/abs/2508.14153}
\end{links}

\section{Introduction}
\label{sec:intro}
Text-prompted segmentation takes a natural-language description and an image as input and returns a fine-grained segmentation mask.
Unlike conventional semantic segmentation that relies on a fixed set of category labels, text-prompted segmentation must jointly interpret language and vision to localize arbitrary open-vocabulary objects.
This requirement gives rise to three core challenges:
(i) cross-modal localization of text-referenced objects, 
(ii) multi-step relational reasoning across modalities,
and (iii) pixel-level alignment between linguistic cues and image regions.
Because of these capabilities, text-prompted segmentation is well suited to real-world scenarios such as robotics, where an agent must understand its environment before acting.

Recent studies \cite{lai2024lisa,bai2024onetokenseg,ren2024pixellm} incorporate multimodal large language models (MLLMs) to improve cross-modal localization.
These methods drive the segmentation process with a single token “\texttt{<seg>}” and train the entire pipeline via supervised fine-tuning (SFT).
However, this paradigm faces two major limitations: 
(i) it neglects the intermediate reasoning process that is essential for complex, reasoning-intensive tasks,
and (ii) its heavy reliance on SFT often leads to overfitting and weak generalization. 
These issues motivate us to pursue a more robust and generalizable test-time reasoning framework.

\begin{figure}[t]
  \centering
  \includegraphics[width=\linewidth]{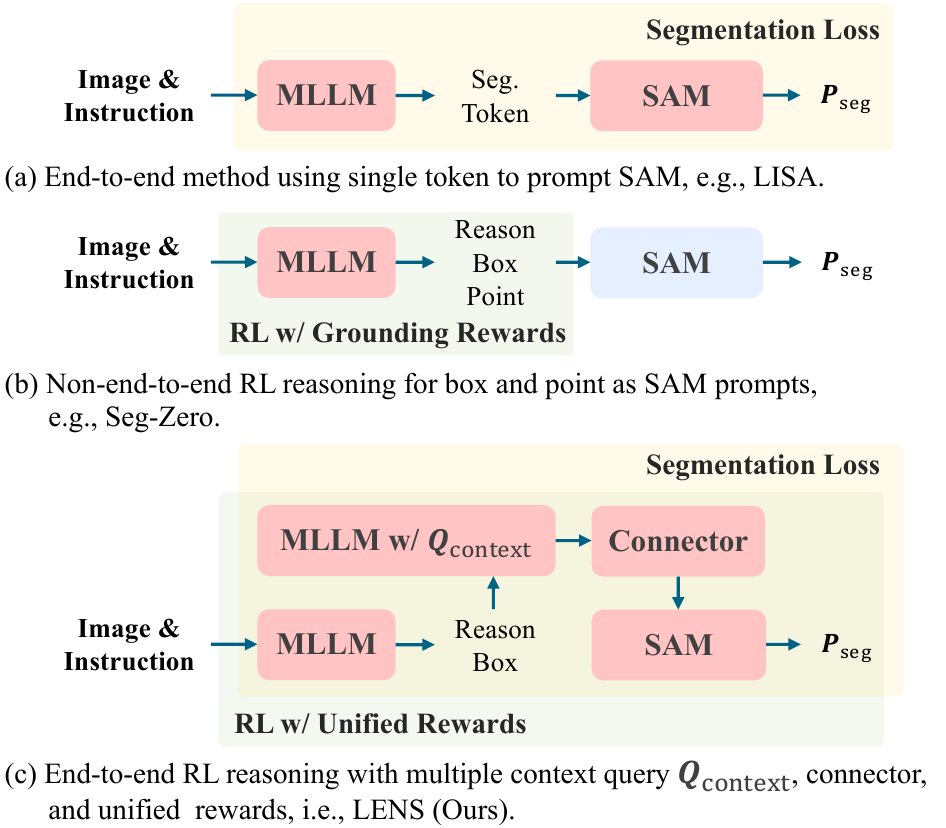}
  \caption{Framework Comparison between proposed \name{} and other methods.}
  \label{fig:teaser}
\end{figure}

Group Relative Policy Optimization (GRPO)~\cite{shao2024deepseekmath} is a rule-based reinforcement-learning (RL) algorithm for post-training large language models. 
By optimizing policies with group-relative rewards, GRPO strengthens reasoning ability and empirically generalizes better than SFT.
Motivated by GRPO, we present \name, a unified, test-time reasoning framework for text-prompted segmentation. 
\name{} adopts the GRPO strategy with the proposed unified rewards that simultaneously consider sentence-, box-, and segment-level cues.
Furthermore, we introduce a context module to bridge the MLLM and the segmentation model, which extracts the reasoning and grounding information as prior to guide the segmentation process.

Concurrent work SegZero~\cite{liu2025segzero} directly feeds the grounding output, i.e., the bounding box and points, of an MLLM into a frozen SAM, achieving strong reasoning segmentation ability.
However, the final mask quality is fully determined by the MLLM’s grounding, and any errors in grounding will propagate downstream, leading to segmentation errors.
In contrast, \name{} offers an end-to-end solution that jointly optimizes language understanding and pixel-wise mask prediction (Fig.~\ref{fig:teaser}).
We establish a tight coupling between the MLLM and SAM through the proposed context module and a pre-alignment stage.
Moreover, the proposed unified reinforcement learning rewards, i.e., format reward, box-IoU reward, and segment-IoU reward, encourage the model to produce informative chain-of-thought rationales while refining mask quality.
Extensive experiments show that \name\ sets a new state-of-the-art on standard text-prompted segmentation benchmarks.

To sum up, our contributions are as follows:
\begin{itemize}
    \item We propose \name{}, an end-to-end, test-time reasoning framework that jointly optimizes multimodal language understanding and pixel-level segmentation for text-prompted tasks.
    \item We introduce a context module, i.e., multiple context queries and a connector, to bridge the MLLM and the segmentation model. Through the pre-alignment stage, the context module can transform the chain-of-thought reasoning trace and grounding box into spatial priors that guide mask generation
    \item Built upon GRPO, we propose a unified rewards scheme that simultaneously supervises sentence-level reasoning, object localization, and pixel-wise mask quality within a single reinforcement-learning objective.  
    \item \name{} achieves 81.2\% average cIoU on RefCOCO-series benchmarks, 58.0\% cIoU on ReasonSeg-Test, and 78.3\% cIoU on GS-Eval, establishing new state-of-the-art performance for text-prompted segmentation. 
\end{itemize}

\section{Related Work}
\label{sec:relatedw}

\paragraph{Text-prompted Segmentation}

Text-prompted segmentation, i.e., referring segmentation, aims to segment objects described by natural language expressions. This task requires both visual understanding and language comprehension capabilities. Early work includes ReferIt~\cite{kazemzadeh2014referitgame} dataset and corresponding models that localize objects based on natural language descriptions.

Recent advances have focused on multimodal fusion architectures. LAVT~\cite{yang2022lavt} proposed a lightweight vision-transformer approach for referring segmentation. RefTR~\cite{li2021referring} employed transformer-based cross-modal fusion. CRIS~\cite{wang2022cris} introduced contrastive learning for improved text-image alignment. More recent work includes X-Decoder~\cite{zou2023generalized} and SEEM~\cite{zou2023segment}, which unify various segmentation tasks including text-prompted segmentation.

The integration of large language models has opened new possibilities. LISA~\cite{lai2024lisa} combines SAM with large language models for reasoning segmentation. PerSAM~\cite{zhang2023personalize} enables few-shot personalization of SAM. However, these approaches primarily rely on supervised fine-tuning, which may limit their reasoning capabilities and generalization to unseen scenarios.

\paragraph{Reinforcement Learning in Large Language Models}

Reinforcement Learning from Human Feedback (RLHF) has become a cornerstone technique for aligning large language models with human preferences. PPO (Proximal Policy Optimization)~\cite{schulman2017proximal} serves as the foundational algorithm, enabling stable policy updates while preventing catastrophic forgetting.

Recent breakthroughs include InstructGPT~\cite{ouyang2022training} and ChatGPT, which demonstrated the effectiveness of RLHF in producing helpful and harmless responses. Constitutional AI~\cite{bai2022constitutional} proposed training models to follow a set of principles. DPO (Direct Preference Optimization)~\cite{rafailov2023direct} simplified the RLHF pipeline by directly optimizing on preference data.

The emergence of reasoning-focused RL approaches has shown remarkable success. DeepSeek-R1~\cite{guo2025deepseek} demonstrated that large-scale reinforcement learning can significantly improve reasoning capabilities, achieving competitive performance with leading models. Chain-of-thought reasoning with RL supervision has been explored in various contexts~\cite{zelikman2022star,huang2022large}, showing that explicit reasoning processes can be learned and improved through reinforcement learning.

However, most existing RL approaches focus on text generation tasks. The application of RL to multimodal tasks, particularly vision-language understanding and segmentation, remains largely unexplored. Our work bridges this gap by introducing RL-based reasoning for text-prompted segmentation, enabling more robust and generalizable segmentation models.

\section{\name}
\label{sec:method}

In this section, we introduce the proposed \name{} architecture and the proposed reinforcement learning training strategy.
First, we introduce the proposed \name{} architecture that enables end-to-end reasoning and segmentation.
Then, we introduce the proposed pretraining alignment stage that establishes the foundational connection between the MLLM and SAM.
Finally, we introduce the proposed reinforcement learning stage that jointly optimizes the model's reasoning and segmentation capabilities.

\begin{figure*}[ht]
  \centering
  \includegraphics[width=.95\textwidth]{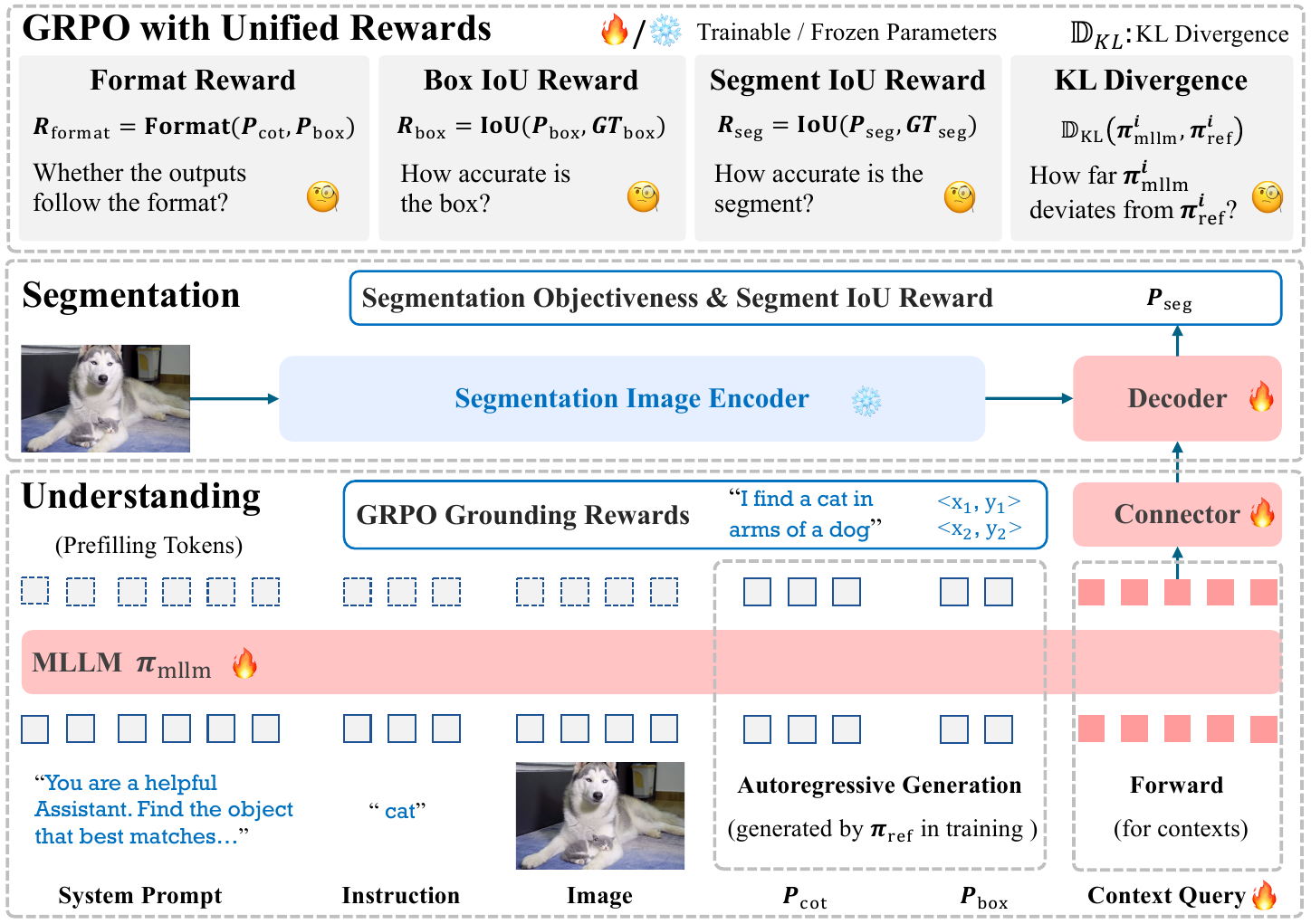}
  \caption{An Overview of \name{} framework. In the pretraining alignment stage, we only train the context query and connector with the segmentation objectiveness. In the reinforcement learning stage, we train all the parts except for the segmentation image encoder with the multi-grained objectiveness, i.e., the unified GRPO rewards and segmentation loss.}
\end{figure*}

\subsection{\name{} Architecture}\label{sec:arch}

The concurrent SegZero~\cite{liu2025segzero} triggers a frozen SAM with bounding boxes and points predicted by an MLLM,
which achieves strong reasoning ability. 
However, its non-end-to-end nature hinders the joint alignment of understanding and segmentation representations and ultimately yields sub-optimal segmentation masks.
To overcome this limitation, we introduce \name, an end-to-end reinforcement-learning framework that couples an MLLM with SAM through a context module.
Thereby, the model can jointly optimize the reasoning and segmentation capabilities under the multi-grained unified rewards and pixel-level entropy loss.

\paragraph{Reasoning Model.}

We use MLLMs, i.e., Qwen2.5-VL~\cite{bai2025qwen2p5vl}, as the reasoning model $\pi_{\text{mllm}}$.
Previous MLLMs~\cite{liu2023llava,bai2023qwenvlversatilevisionlanguagemodel} show promising performance in object-centric localization, but still lack pixel-level perception ability.
Some work~\cite{lai2024lisa,bai2024onetokenseg} incorporates pre-trained segmentation models into MLLMs to compensate the lack of pixel-level perception ability.
However, these methods simply use a segmentation token in short response, i.e., `It is $\mathtt{<seg>}$', to guide the segmentation, which have neglected the reasoning process, e.g., Chain-of-Thought (CoT)~\cite{wei2022cot}.
To address this, we propose to utilize the reasoning process to guide the segmentation, which brings more robust and effective segmentation ability.
We input the system prompt $p_{\text{sys}}$, textual instruction $T$ and image $I$ to the reasoning model, and obtain the CoT $P_{\text{CoT}}$ and box prediction $P_{\text{box}}$.
\begin{align}
  P_{\text{CoT}}, P_{\text{box}} = \pi_{\text{mllm}}.\mathtt{generate}(p_{\text{sys}}, T, I).
\end{align}

\paragraph{Context Module.}

Concurrent method~\cite{liu2025segzero} only rely detokenized bounding box and points to prompt the SAM, which leads to suboptimal performance and eliminates the end-to-end optimization.
To address this, we propose a context module to bridge the gap between understanding and segmentation during the reinforcement learning process.
The context module contains the context query and the connector. 
After the MLLM generates the CoT and box prediction, we use the context query to extract the context information from the MLLM.
Then, we use the connector to project the context information as the prompt.

We randomly initialize the context query $Q \in \mathbb{R}^{M \times C}$, where $M$ is the number of context queries and $C$ is the hidden dimension of the MLLM.
Notably, the context query is appended to the end of the input, and generated sequence, gathering the information through a forward pass.
\begin{align}
Q' &= \pi_{\text{mllm}}.\mathtt{forward}(p_{\text{sys}}, T, I, P_{\text{CoT}}, P_{\text{box}}, Q), \\
Q&_{\text{seg}} = \pi_{\text{connector}}(Q')
\end{align}

Next, we feed the context query embeddings $Q' \in \mathbb{R}^{M \times C}$ output by the MLLM to the connector, which is a shallow transformer to project the context query embeddings to the prompt space of the SAM, i.e., $Q_{\text{seg}}$.

\paragraph{Segmentation Model.}
SAM~\cite{kirillov2023sam} is the most popular segmentation model~\cite{cheng2025cross}, which accepts prompts including point and box and produces precise segmentation results.
Some methods~\cite{zhang2024evfsam,lai2024lisa} incorporate MLLMs, which enable SAM to segment objects with text prompts.
However, these methods output the single segmentation token `$\mathtt{<seg>}$' directly, which neglects the reasoning process and token capacity.
To address this, we propose to use multiple queries to extract reasoning context and ensure the enough token capacity.
Moreover, we incorporate pixel-level entropy loss to reinforcement learning training, which provides more fine-grained optimization guidance.
We feed the projected segmentation prompt $Q_{\text{seg}}$ to the SAM model $\pi_{\text{sam}}$ for segmentation $P_{\text{seg}}$.
\begin{align}
  P_{\text{seg}} = \pi_{\text{sam}}(Q_{\text{seg}}, I).
\end{align}

\subsection{Pretraining Alignment Stage}\label{sec:pretrain}

In this stage, we establish the foundational connection between the MLLM and SAM to enable end-to-end segmentation. Since understanding and segmentation models have different pretrained representations, we introduce the context module to bridge this gap while preserving the original capabilities of both models.

During pretraining alignment, we freeze the weights of both the MLLM ($\pi_{\text{mllm}}$) and SAM ($\pi_{\text{sam}}$), training only the lightweight context module, i.e., the connector and the context queries. This approach ensures that the rich pretrained knowledge in both models is preserved while enabling them to work together effectively.

The training objective focuses on learning segmentation-aware representations through the context query and task connector. 
We train the model on segmentation datasets where the context query learns to extract relevant visual-semantic information from the MLLM's representations, and the task connector learns to transform this information into effective prompts for SAM.

We use segmentation objectives in this stage:
\begin{align}
  \mathcal{L}_{\text{align}} = \mathcal{L}_{\text{seg}}(P_{\text{seg}}, M_{\text{gt}}),
\end{align}
where $M_{\text{gt}}$ represents the ground truth segmentation mask and $\mathcal{L}_{\text{seg}}$ is the segmentation loss (e.g., dice loss, focal loss).

\subsection{Reinforcement Learning Stage}\label{sec:rl}

While the pretraining alignment stage enables basic multimodal understanding and segmentation, the quality of reasoning process significantly impacts segmentation performance. 
Existing methods~\cite{liu2025segzero,shen2025vlmr1} typically optimize only the understanding part, overlooking the crucial segmentation decoder. 

To address this limitation, we propose unified rewards ranging from sentence-, box-, and segment-level to jointly optimize the reasoning quality and segmentation accuracy.
In this stage, we unfreeze the MLLM and segmentation decoder parameters while keeping segmentation image encoder frozen, allowing the model to learn enhanced reasoning strategies. 
The improved Chain-of-Thought (CoT) reasoning $P_{\text{CoT}}$ serves as richer contextual information to strengthen the context query representations, ultimately leading to better segmentation results.

\paragraph{Group Relative Policy Optimization (GRPO).} 
The GRPO~\cite{shao2024deepseekmath} introduces a new rule-based reinforcement learning algorithm for post-training optimization of large language models. 
Traditional reinforcement learning algorithms, i.e., PPO~\cite{schulman2017ppo}, use separate critic models to evaluate the quality of the policy model, while leading to high memory and computational costs.
To alleviate this problem, GRPO eliminates the separate critic model and only uses group relative rewards to guide the optimization of the policy model. 
GRPO employs reward function $R(\cdot)$ to evaluate the outputs and Kullback-Leibler (KL) divergence penalty $\mathbb{D}_{\text{KL}}$ to regularize the policy update.
Due to the page limitation, we leave more details in the appendix.

\begin{table*}[ht]
    \centering    
    \begin{tabular}{lccccccccc}\toprule
        \multirow{2}{*}{Method}& \multicolumn{3}{c}{RefCOCO} & \multicolumn{3}{c}{RefCOCO+} & \multicolumn{2}{c}{RefCOCOg} &  \multirow{2}{*}{AVG} \\
         &val& testA & testB & val & testA & testB & val & test & \\\midrule
         \multicolumn{6}{l}{\textbf{\textit{without active chain-of-thought (CoT) reasoning}}}\\
         LAVT~\cite{yang2022lavt}(CVPR 22) & 72.7 & 75.8 & 68.8 & 62.1 & 68.4 & 55.1 & 61.2 & 62.1 & 65.8\\
         ReLA~\cite{liu2023gres}(CVPR 23) & 73.8 & 76.5 & 70.2 & 66.0 & 71.0 & 57.7 & 65.0 & 66.0 & 68.3\\
         X-Decoder~\cite{zou2023generalized}(CVPR 23) & - & - &- &- &- &- & 64.6 &- & - \\
         SEEM~\cite{zou2023segment}(NeurIPS 23) &- &- &- &- &- &- & 65.7 &- & - \\
         LISA~\cite{lai2024lisa} (CVPR 24) & 74.1 & 76.5  & 71.1  & 62.4 &	67.4  &	56.5  & 66.4 & 68.5 & 67.9 \\
         PixelLM~\cite{ren2024pixellm}(CVPR 24) & 76.9 & 78.5 & 74.4 & 69.2 & 72.1 & 64.5 & 70.7 & 72.4 & 72.3\\
         PerceptionGPT-7B~\cite{pi2024perceptiongpt}(CVPR 24) & 75.1 & 78.6 & 71.7 & 68.5 & 73.9 & 61.3 & 70.3 & 71.7 & 71.4\\
         OMG-LLaVA~\cite{zhang2024omg}(NeurIPS 24) & 78.0 & 80.3 & 74.1 & 69.1 & 73.1 & 63.0 & 72.9 & 72.9 & 72.9 \\
         VISA~\cite{yan2024visa}(ECCV 24) & 72.4 & 75.5 & 68.1 & 59.8 & 64.8 & 53.1 & 65.5 & 66.4 & 65.7\\
         GLaMM~\cite{rasheed2024glamm}(CVPR 24) & 79.5 & \underline{83.2}  & 76.9  & 72.6 & \underline{78.7}  &	64.6  &	74.2 & 74.9 & 75.6\\
         SAM3-Agent-Gemini2.5-Pro (OpenReview 2025) & 74.9 & 77.8 & 69.9 & 66.9 & 71.1 & 62.4 & 73.3 & 73.6 & 74.2\\
         \midrule
         \multicolumn{6}{l}{\textbf{\textit{with active chain-of-thought (CoT) reasoning}}}\\
         Seg-Zero-3B~\cite{liu2025segzero} (arXiv 25) & - & 79.3 & - & - & 73.7 & - & 71.5 &- &-\\
         Seg-Zero-7B~\cite{liu2025segzero} (arXiv 25) & - & 80.3 & - & - & 76.2 & - & 72.6 & -&-\\         
         \name{}-2B (Ours) & \underline{80.7} & 82.7 & \underline{77.1} & \underline{73.8} & 78.0 & \underline{67.3} & \underline{75.7} & \underline{76.4} & \underline{76.5}\\
         \name{}-3B (Ours) & \textbf{84.2} & \textbf{85.3} & \textbf{81.0} & \textbf{79.4} & \textbf{82.8} & \textbf{74.3} & \textbf{81.2} & \textbf{81.0} & \textbf{81.2}\\
         \bottomrule
    \end{tabular}
    \caption{Comparison on the Referring Expression Segmentation (RES) task using cIoU as the evaluation metric. Our method achieves state-of-the-art performance across all RefCOCO, RefCOCO+, and RefCOCOg benchmarks.\textbf{Bold} indicates the best (SoTA) performance, and \underline{underlined} denotes the second-best. The “active chain-of-thought (CoT) reasoning” means using explicit chain-of-thought at test time.}
    \label{tab:refseg_result}
\end{table*}

\paragraph{GRPO with Unified Rewards.}
Reward function is the key to the success of reinforcement learning, as it guides the model to learn the desired behavior.
Text-driven segmentation tasks encompasses both understanding and segmentation, which is more complex than standard MLLM tasks.

Unlike standard VLM-R1 methods~\cite{shen2025vlmr1,yu2025perceptionr1} that relies on purely understanding reward signal, i.e., format reward and box IoU reward, 
we introduce a comprehensive multi-faceted reward system specifically designed for both understanding and segmentation. 
Our reward function encompasses three complementary components:

\begin{itemize}
\item \textbf{Format Reward ($R_{\text{format}}$):} 
Ensures the MLLM output adheres to the expected structure and format consistency. 
The format template requires the MLLM to output the reasoning process in `\textless thinking\textgreater' ... `\textless /thinking\textgreater' tag pair and localization results in `\textless answer\textgreater' ... `\textless /answer\textgreater' tag pair. 
If the format is correct, the reward is 1, otherwise, the reward is 0.
\item \textbf{Box IoU Reward ($R_{\text{box}}$):} 
Measures localization accuracy through IoU between predicted and ground truth bounding boxes. 
The range is $[0, 1]$.
\item \textbf{Segment IoU Reward ($R_{\text{seg}}$):} 
Evaluates overall segmentation quality using mask-level IoU. 
We introduce the segment IoU reward to evaluate the segmentation quality, whose range is $[0, 1]$.

\end{itemize}

Additionally, we use a KL Divergence ($\mathbb{D}_{\text{KL}}$) regularization term to prevent the context query representations from deviating significantly from the pretrained representations.

The unified reward function is formulated as:
\begin{align}
R_{\text{unified}} = \lambda_1 R_{\text{format}} + \lambda_2 R_{\text{box}} + \lambda_3 R_{\text{seg}},
\end{align}
where $\{\lambda_i\}_{i=1}^3$ are balancing hyperparameters.

\paragraph{Training Objective.}
To achieve comprehensive optimization, we combine the unified-reward-based GRPO objective with supervised segmentation loss:
\begin{align}
\mathcal{J}_{\text{\name}}(\theta) &= \mathcal{J}(\theta; R_{\text{unified}}, \mathbb{D}_{\mathrm{KL}})
+ \alpha \mathcal{L}_{\text{seg}}(P_{\text{seg}}, M_{\text{gt}})
\end{align}
where $\mathcal{J}(\theta; R_{\text{unified}}, \mathbb{D}_{\text{KL}})$ represents the standard GRPO objective using our unified rewards and KL divergence regularization, $\alpha$ controls the balance between reinforcement learning and supervised learning signals, and $\mathcal{L}_{\text{seg}}(P_{\text{seg}}, M_{\text{gt}})$ is the segmentation loss.

This joint optimization enables the model to benefit from both reward-driven reasoning improvements and direct segmentation supervision, resulting in enhanced performance across reasoning and segmentation capabilities.

\section{Experiment}
\label{sec:exp}
\paragraph{Implementation Details}
We employ Qwen-VL series models\cite{bai2025qwen2p5vl, wang2024qwen2} as the reasoning model and SAM2-Large~\cite{ravi2024sam2segmentimages} as the segmentation model. 
Specifically, \name-2B is utilizes Qwen2-VL and \name-3B uses Qwen2.5-VL.
Training is conducted on 16 NVIDIA L40S GPUs, with our pipeline built upon the DeepSpeed engine~\cite{rasley2020deepspeed}. 
Detailed experimental settings are summarized in the Appendix.

\paragraph{Benchmarks.} We evaluate the proposed framework on multiple benchmarks: 
Referring Expression Segmentation (RES) benchmarks (RefCOCO series)~\cite{nagaraja2016modeling, yu2016modeling}, 
GroundingSuite benchmarks~\cite{hu2025groundingsuite}, 
and Reasoning Segmentation benchmarks~\cite{lai2024lisa}.
The quantitative comparisons are presented in Table~\ref{tab:refseg_result} and Table~\ref{tab:reasonseg_result}, respectively. 

\paragraph{Evaluation Metrics.}
Following previous works on referring segmentation~\cite{kazemzadeh2014referitgame}, we adopt two commonly used evaluation metrics: generalized IoU (gIoU) and cumulative IoU (cIoU). 
Specifically, gIoU is computed as the average of per-image Intersection-over-Union (IoU) scores, while cIoU is defined as the IoU of the cumulative predicted and ground-truth masks across the entire dataset.

\subsection{Main Results}

\begin{table*}[ht]
    \centering    
    \begin{tabular}{lccccCC}
        \toprule
        \multirow{2}{*}{Method} & \multicolumn{2}{c}{ReasonSeg-Val} & \multicolumn{2}{c}{ReasonSeg-Test} & \multicolumn{2}{c}{GroundingSuite-Eval} \\
         & gIoU & cIoU & gIoU & cIoU & gIoU & cIoU\\    
        \midrule
        PSALM~\cite{zhang2024psalm}(ECCV 24) & - & - & - & - & 39.2 & 34.4\\
        HyperSeg-3B$\ddagger$~\cite{wei2024hyperseg}(CVPR 25) & 59.2 & 56.7 & - & - & 58.2 & 62.8 \\
        InstructSeg-3B$\ddagger$~\cite{wei2024instructseg}(ICCV 25) & 61.9 & \textbf{65.2} & - & - & 55.7 & 57.0 \\
        LISA-7B(ft)~\cite{lai2024lisa} (CVPR 24)           & 52.9 & 54.0 & 55.6 & \underline{56.9} & 60.9 & 68.6 \\
        Seg-Zero-3B~\cite{liu2025segzero} (arXiv 25)           & 58.2 & 53.1 & 56.1 & 48.6 & \textbf{67.3} & \underline{68.9} \\
        SAM3-Agent-Qwen2.5-VL-7B (OpenReview 2025) & \textbf{65.4} & 50.5 & \textbf{62.6} & 56.2 & - & -\\
        \midrule
        \name{}-3B (Ours) & \underline{62.1} & \underline{64.9} & \underline{57.2} & \textbf{58.0} & \underline{67.0} & \textbf{78.3} \\
        \bottomrule
    \end{tabular}
    \caption{Performance comparison on the ReasonSeg and GroundingSuite-Eval benchmarks, evaluated using gIoU and cIoU. \textbf{Bold} and underlined entries indicate the best and second-best results, respectively. $\ddagger$ marks models trained on both train and test splits of ReasonSeg. All models are not trained on GroundingSuite. LENS shows superior reasoning and generalization ability.}
    \label{tab:reasonseg_result}
\end{table*}

\begin{figure*}[h!]
    \centering
    \includegraphics[width=\textwidth]{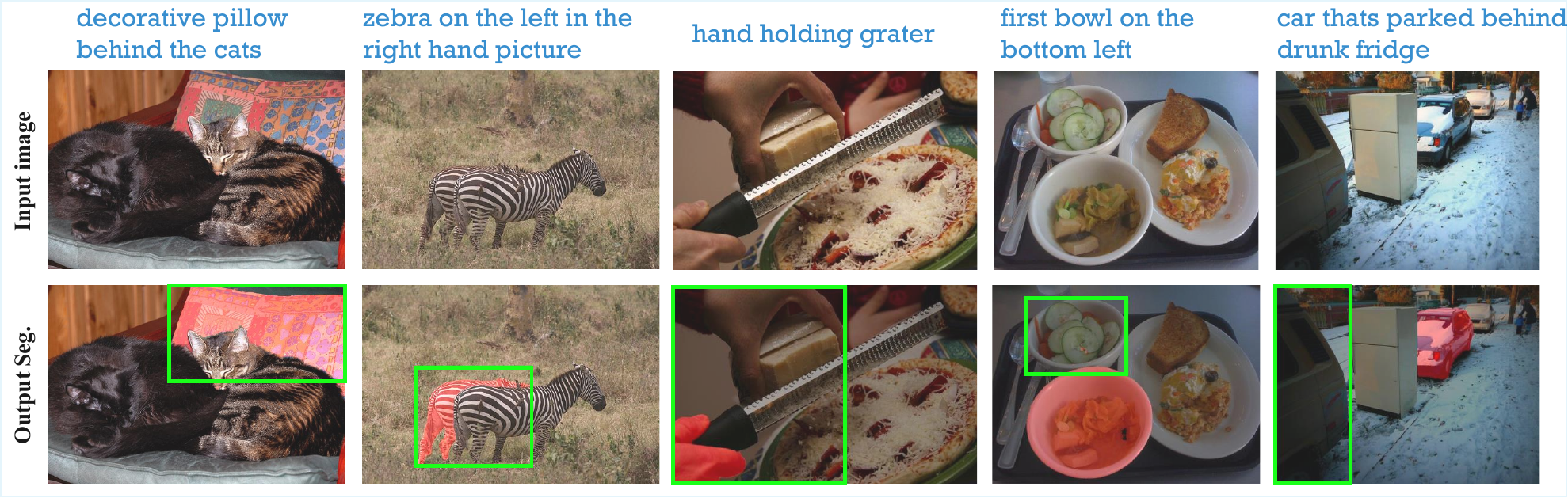}
    \caption{Qualitative results on the Referring Expression Segmentation task. The proposed \name{} can accurately segment partially obscured objects. Benefit from the proposed unified framework, even if there is an error in the context box, the segmentation module can correct it based on the rich context in the multiple queries.}
    \label{fig:refcocog_vis}
\end{figure*}

Note that PSALM~\cite{zhang2024psalm}, HyperSeg~\cite{wei2024hyperseg}, and InstructSeg~\cite{wei2024instructseg} utilize pre-trained Mask2Former~\cite{cheng2022mask2former} models, whose training data, i.e., COCO 2017~\cite{lin2014microsoft}, has overlap with the RefCOCO benchmarks. 
Therefore, we compare with them in Table~\ref{tab:reasonseg_result}, evaluating their performance on the Reasoning Segmentation and GroundingSuite-Eval benchmarks.

\begin{table}[ht]
    \centering
        \centering        
        \begin{tabular}{cccccc}
            \toprule
            Query Num. & 1 & 16 & 32 & 64 & 128 \\
            \midrule
            cIoU (\%) & 85.9 & 87.4 & 87.8 & \textbf{87.9} & 87.2 \\
            \bottomrule
        \end{tabular}
        \caption{Ablation study on the number of queries used in context module, evaluated cIoU metric on RefCOCO testA in pretraining alignment stage. Performance improves with more queries up to 64, after which it slightly drops, indicating a saturation point.}
        \label{tab:num of query ablation}
\end{table}

\begin{table}[ht]
    \centering
        \centering        
        \begin{tabular}{ccc}
            \toprule
            Connector & MLP & ViT \\
            \midrule
            cIoU (\%) & 72.8 & 87.8 \\
            \bottomrule
        \end{tabular}
        \caption{Ablation study on the connector module between MLLM and segmentation model, evaluated cIoU metric on RefCOCO testA in pretraining alignment stage. Using a ViT connector significantly outperforms a simple MLP, demonstrating the benefit of attention-based context fusion.}
        \label{tab:connector ablation}
\end{table}

\begin{table}[ht]
\centering
\begin{tabular}{ccccc c}
\toprule
\multirow{2}{*}{\raisebox{-0.7\totalheight}{\shortstack{Context\\Module}}} & 
\multirow{2}{*}{\raisebox{-0.65\totalheight}{\shortstack{Align.\\Stage}}} & \multicolumn{3}{c}{RL Stage} & \multirow{2}{*}{cIoU} \\
\cmidrule(lr){3-5}
& & $R_{\text{box}}$ & $R_{\text{format}}$ & $R_{\text{seg}}$ & \\
\midrule
$\times$ & $\times$ & $\times$ & $\times$ & $\times$ & 79.3 \\
$\checkmark$ & $\times$ & $\checkmark$ & $\times$ & $\times$ & -- \\
$\checkmark$ & $\checkmark$ & $\times$ & $\times$ & $\times$ & 80.9 \\
$\checkmark$ & $\times$ & $\checkmark$ & $\checkmark$ & $\times$ & 72.3 \\
$\checkmark$ & $\checkmark$ & $\checkmark$ & $\checkmark$ & $\times$ & 81.9 \\
$\checkmark$ & $\checkmark$ & $\checkmark$ & $\checkmark$ & $\checkmark$ & \textbf{82.7} \\
\bottomrule
\end{tabular}
\caption{Ablation study of components on RefCOCO testA.}
\label{tab:ablation-context-rl}
\end{table}

\paragraph{Referring Expression Segmentation.}
As shown in Table~\ref{tab:refseg_result}, the proposed \name{} achieves state-of-the-art performance across all splits of the RefCOCO, RefCOCO+, and RefCOCOg benchmarks. 
Compared to prior models such as GlaMM~\cite{rasheed2024glamm}, our method delivers significant improvements, especially on RefCOCO+ and RefCOCOg, which are known to require a more precise understanding of compositional and spatial language. 
Notably, \name-3B surpasses the best previous method by 6.1\% on RefCOCOg-test, highlighting the model's strength in handling complex referring expressions. 

\paragraph{ReasonSeg and GroundingSuite-Eval Benchmarks.}
As shown in Table~\ref{tab:reasonseg_result}, we evaluate the proposed \name{} on the ReasonSeg and GroundingSuite-Eval benchmarks. GroundingSuite (GS) is designed for challenging scenarios, and none of the methods in Table~\ref{tab:reasonseg_result} are trained on GS-Train. On the zero-shot GS-Eval benchmark, \name{} achieves a 9.4\% improvement over the second-best method, demonstrating strong out-of-domain (OOD) generalization.

In terms of segmentation quality, \name{} achieves competitive results in gIoU and exhibits a clear lead in cIoU, which captures holistic segmentation performance across objects and scenes. For example, on GS-Eval, \name{} attains a cIoU of 78.3\%, surpassing Seg-Zero-3B by 9.4\%.

\subsection{Ablation Studies}

\begin{figure*}[ht!]
    \centering
    \includegraphics[width=\textwidth]{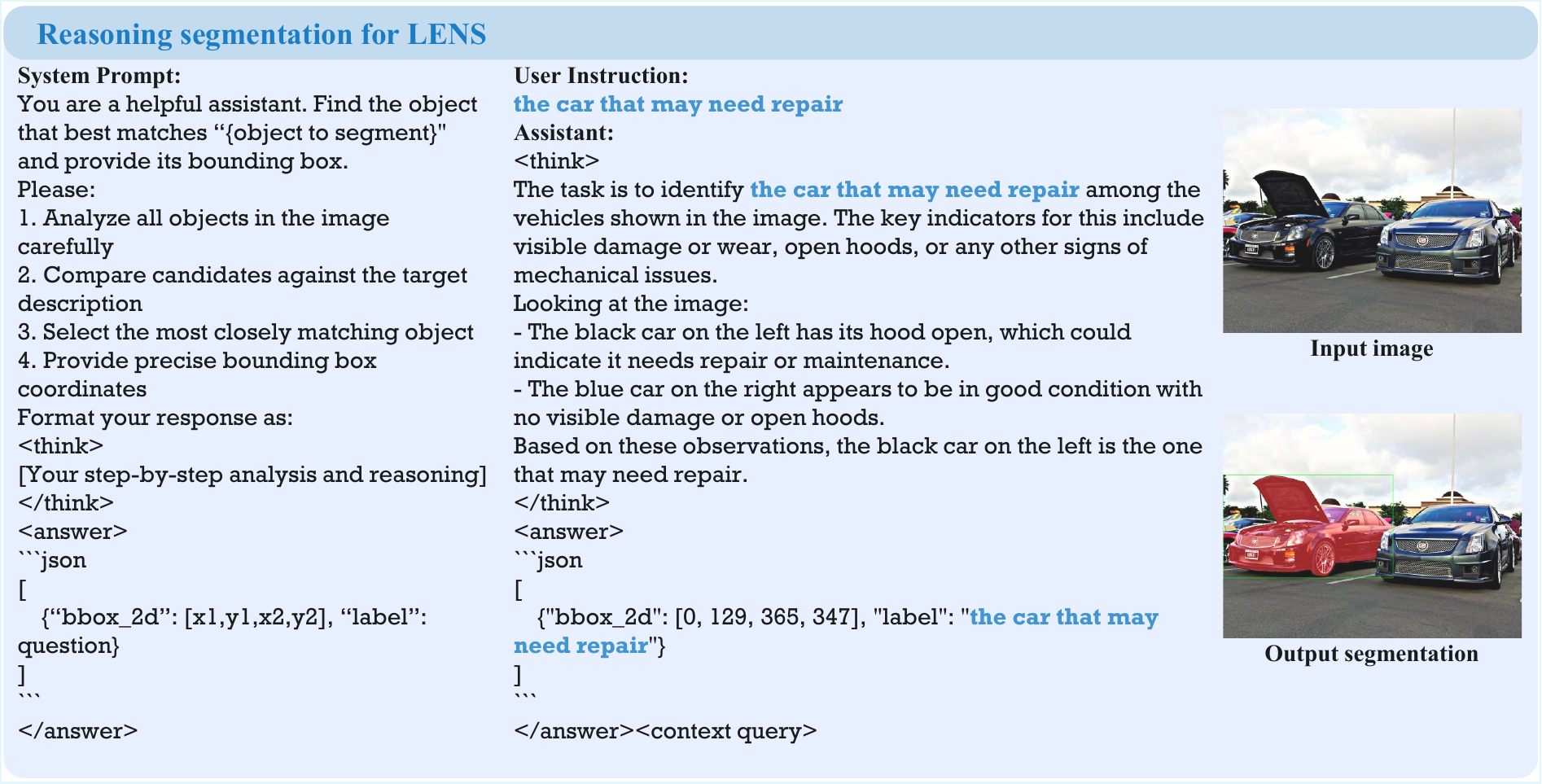}
    \caption{Visualization on reasoning segmentation. MLLM first generates a CoT reasoning process and a probable box as priors. Then the context queries extract messages from priors, and prompt segmentation module for an accurate mask.}
    \label{fig:reasonseg}
\end{figure*}

\paragraph{Number of Queries.}
We examine the impact of varying the number of queries in the reasoning module. 
As shown in Table~\ref{tab:num of query ablation}, performance improves steadily from 1 to 64 queries, suggesting that a larger query set enables the model to capture more contextual reasoning cues and deliver to the segmentation task. 
However, using 128 queries leads to a slight drop, likely due to increased redundancy and computational overhead. 
This indicates that 64 queries offer a good trade-off between accuracy and efficiency.

\paragraph{Connector Design.}
We further investigate the role of the proposed connector bridging the MLLM and the segmentation head. 
The connector is implemented as a 4-layer Qwen-2.5 Transformer with a hidden dimension of 2048, matching the MLLM output. 
As shown in Table~\ref{tab:connector ablation}, replacing a simple MLP with this Transformer significantly improves performance. 
Overall, these studies underscore the importance of architectural choices in enabling effective reasoning and accurate segmentation within the proposed unified framework.

\paragraph{Framework Analysis.}
To investigate the contribution of each component in our framework, we conduct an ablation study on the RefCOCO testA split, as shown in Table~\ref{tab:ablation-context-rl}. 
We start with a Lisa-liked baseline~\cite{lai2024lisa} model without the context module, alignment stage, or reinforcement learning (RL), which achieves a cIoU of 79.3\%. 
Directly adding the context module and applying the box reward lead to instable.
After enabling format reward, the model captures basic localization cues, but the lack of explicit alignment between textual and visual features leads to sub-optimal performance.
To address this issue, we enable the alignment stage, where the context module is pretrained to align MLLM and SAM. 
This significantly improves the segmentation accuracy, boosting cIoU to 81.9\%, and demonstrates the critical importance of semantic alignment in the proposed unified framework. 
Finally, introducing the segmentation-level reward ($R_{\text{seg}}$) further refines the mask quality, pushing the performance to 82.7\% cIoU.

\paragraph{Visualizations}
We provide qualitative results to illustrate the effectiveness of our framework on both general and reasoning-intensive segmentation tasks.
Fig.~\ref{fig:refcocog_vis} shows examples from the RefCOCOg benchmark. \name{} accurately segments target objects based on referring expressions, demonstrating strong performance in scenes with partially obscure objects and multiple similar instances.
Notably, the proposed unified framework and end-to-end optimization make \name{} robust to correct the potential error in the box prior.
Fig.~\ref{fig:reasonseg} presents results on the reasoning segmentation benchmark. \name{} exhibits robust test-time reasoning capabilities, effectively handling tasks that require spatial understanding, multi-step inference, and comparative reasoning.
These visualizations highlight the \name’s ability to integrate high-level textual reasoning with fine-grained visual perception. 

\section{Conclusion}
\label{sec:conclu}
In this paper, we introduce \name{}, a unified test-time reasoning framework for text-prompted segmentation.
By coupling a multimodal LLM with a segmentation model through a context module and a pretraining alignment stage, \name{} treats language understanding and mask prediction as equally critical components.
The extracted chain-of-thought rationales serve as spatial priors that directly guide segmentation. 
In addition, we design a unified reinforcement-learning (RL) objective with unified rewards that simultaneously supervise sentence-level reasoning, object localization, and pixel-accurate masking.
We believe \name{} offers fresh insights into the seamless integration of RL and visual segmentation and will spur further research toward more general, robust, and intelligent vision–language systems.

\textbf{Acknowledgement}: This work was partially supported by the National Natural Science Foundation of China (No. 62276108).

\bibliography{samr1}

\clearpage %
\appendix
\section{Appendices}
\label{sec:appendix}
\renewcommand{\thefigure}{A\arabic{figure}} 
\renewcommand{\thetable}{A\arabic{table}} 
\setcounter{figure}{0}
\setcounter{table}{0}
\subsection{More GRPO Details}

Given a question sample $q$, GRPO first samples a group of outputs $\{o_i\}_{i=1}^G$ with the policy model $\pi$, where $G$ is the number of samples. 
Next, it uses reward function $\mathtt{R}(\cdot)$ to calculate the rewards for each output.
To determine the relative quality of the outputs in the group, GRPO then calculates the advantage $A_i$ for each output as follows:
\begin{align}
\mu_G &= \frac{1}{G} \sum_{i=1}^G \mathtt{R}(o_i), \sigma_G = \sqrt{\frac{1}{G} \sum_{i=1}^G \left(\mathtt{R}(o_i) - \mu_G\right)^2}, \\
A_i &= (\mathtt{R}(o_i) - \mu_G) / (\sigma_G + \delta),
\end{align}
where $\delta$ is a small constant to avoid division by zero.  
Subsequently, GRPO estimate clipped surrogate objective $\mathcal{O}_{\text{clip}}^{i}$ and Kullback-Leibler (KL) divergence penalty $\mathbb{D}_{\text{KL}}[\pi_\theta || \pi_{\text{ref}}]$ following the style of PPO:
\begin{align}
  \mathcal{O}_{\text{clip}}^{i} &= \min \left[ \frac{\pi_\theta^{i}}{\pi_{\theta_{\text{old}}}^{i}} A_{i}, \text{clip} \left( \frac{\pi_\theta^{i}}{\pi_{\theta_{\text{old}}}^{i}}, 1-\epsilon, 1+\epsilon \right) A_{i} \right],\\
  \mathbb{D}_{\text{KL}}&[\pi_\theta || \pi_{\text{ref}}] = \frac{\pi_{\text{ref}}^i}{\pi_\theta^i} - \log \frac{\pi_{\text{ref}}^i}{\pi_\theta^i} - 1,
\end{align}
where $\pi_{\theta}^i$, $\pi_{\text{old}}^i$, and $\pi_{\text{ref}}^i$ represent the probability of the $i$-th output under the policy, old policy, and reference models, respectively. 
$\epsilon$ is the clipping range parameter.
Finally, GRPO optimizes the policy model with the following objective function:
\begin{align}
\mathcal{J}(\theta) 
&= \mathbb{E}_{\substack{\{o_i\}_{i=1}^G }} \Bigg[ %
\frac{1}{G} \sum_{i=1}^G \left( \mathcal{O}_{\text{clip}}^i 
- \beta\, \mathbb{D}_{\text{KL}}[\pi_\theta^i \,||\, \pi_{\text{ref}}^i] \right) \Bigg],
\end{align}
where $\beta$ is the coefficient of the KL divergence penalty.

\subsection{More Implementation Details}

In this section, we summarize the training and evaluation
settings for reproduction.

\subsubsection{Pretraining Alignment Stage}

For the pretraining alignment stage, 
we follow the settings in SAM2~\cite{ravi2024sam2segmentimages} and Qwen2.5-VL~\cite{bai2025qwen2p5vl}.
Moreover, we list the important experimental settings as shown in Table~\ref{tab:stage1-exp-settings}. 
Note that the training set in this stage is from the RefCOCO series~\cite{nagaraja2016modeling, yu2016modeling}, which includes RefCOCO, RefCOCO+, and RefCOCOg.

\begin{table}[t]
\centering
\begin{tabular}{l|l}
\toprule
\textbf{config} & \\
\midrule
epochs & 25 \\
batch size & 128 \\
SAM2 image size & $1024^2$ \\
MLLM image size & default smart resize \\
learning rate & $3 \times 10^{-5}$ \\
scheduler & Cosine \\
seed & 42 \\
number of context query & 64 \\
max prompt length & 2048\\
max completion length & 768\\
\bottomrule
\end{tabular}
\caption{Pretraining alignment stage experimental settings. We use AdamW~\cite{loshchilov2017decoupled} optimizer.}
\label{tab:stage1-exp-settings}
\end{table}

\subsubsection{Reinforcement Learning Stage}

For the referring expression segmentation, we follow the experimental settings in VLM-R1~\cite{shen2025vlmr1} as shown in Table~\ref{tab:stage2-exp-settings}.

\begin{table}[t]
\centering
\begin{tabular}{l|l}
\toprule
\textbf{config} & \\ %
\midrule
epochs & 16 \\
batch size & 64 \\
SAM2 image size & $1024^2$ \\
MLLM image size & default smart resize \\
learning rate & $3 \times 10^{-6}$ \\
scheduler & Linear \\
seed & 42 \\
number of context query & 64 \\
reward weight $\lambda_{i}$ & 1.0 \\
segmentation weight $\alpha$ & 1.0 \\
number of GRPO samples $G$ & 8 \\
max prompt length & 2048\\
max completion length & 768\\
\bottomrule
\end{tabular}
\caption{Reinforcement learning stage experimental settings. Optimizer is AdamW~\cite{loshchilov2017decoupled}.}
\label{tab:stage2-exp-settings}
\end{table}

For the referring expression segmentation, our training set is from the RefCOCO series, including RefCOCO, RefCOCO+, and RefCOCOg. For the reasoning segmentation, we adopt the ReasonSeg~\cite{lai2024lisa} for training.
\subsection{More Ablations}
\begin{table}[ht]
\centering
\begin{adjustbox}{max width=\linewidth}
\begin{tabular}{cccccc}
\toprule
\makecell[c]{SAM\\Version} & \makecell[c]{Tune\\SAM} & \makecell[c]{Context\\Module} & \makecell[c]{Align.\\Stage} & \makecell[c]{RL\\Stage} & \makecell[c]{cIoU} \\
\midrule
1 & $\times$ & $\times$ & $\times$ & $\times$ & 78.5 \\
1 & $\checkmark$ & $\times$ & $\times$ & $\times$ & 79.1 \\
2 & $\checkmark$ & $\times$ & $\times$ & $\times$ & 79.3 \\
2 & $\checkmark$ & $\checkmark$ & $\times$ & $\times$ & 80.4 \\
2 & $\checkmark$ & $\checkmark$ & $\checkmark$ & $\times$ & 80.9 \\
2 & $\checkmark$ & $\checkmark$ & $\checkmark$ & $\checkmark$ & \textbf{82.7} \\
\bottomrule
\end{tabular}
\end{adjustbox}
\caption{Extended ablation study on SAM variants and training strategy on RefCOCO testA.}
\label{tab:ablation-context-rl-supp}
\end{table}
\begin{table}[ht]
\centering
\begin{adjustbox}{max width=\linewidth}
\begin{tabular}{cccccc}
\toprule
\makecell[c]{ } & \makecell[c]{MLLM} & \makecell[c]{Context\\Query} & \makecell[c]{Connector} & \makecell[c]{SAM2} \\
\midrule
Params (M) & 3755.6 & 0.13 & 206.3 & 224.4 \\
Latency (s) & 1.5 & 0.07 & 0.002 & 0.025 \\
\bottomrule
\end{tabular}
\end{adjustbox}
\caption{Parameter and latency of model components.}
\label{tab:module_details}
\end{table}
We conduct an additional ablation on RefCOCO testA to examine the impact of SAM variants and training choices (Table~\ref{tab:ablation-context-rl-supp}). Replacing SAM2 by SAM1 or freezing the mask decoder reduces performance, which highlights the benefit of a stronger SAM backbone and fine-tuning the mask decoder. Adding the context module improves cIoU to 80.4\% by capturing sentence-level and cross-modal cues. Introducing the Alignment Stage, which explicitly aligns textual and visual features, further increases performance to 80.9\%, demonstrating the importance of semantic alignment between the MLLM and SAM. Finally, incorporating the segmentation-level RL stage refines mask quality and yields the largest gain, reaching 82.7\% cIoU. Overall, these results indicate that a stronger off-the-shelf segmenter, context modeling, explicit alignment, and RL each provide complementary benefits and collectively lead to substantial improvements.

\subsection{Rewards and Loss During Training}
\begin{figure*}[ht]
    \centering
    \includegraphics[width=.95\textwidth]{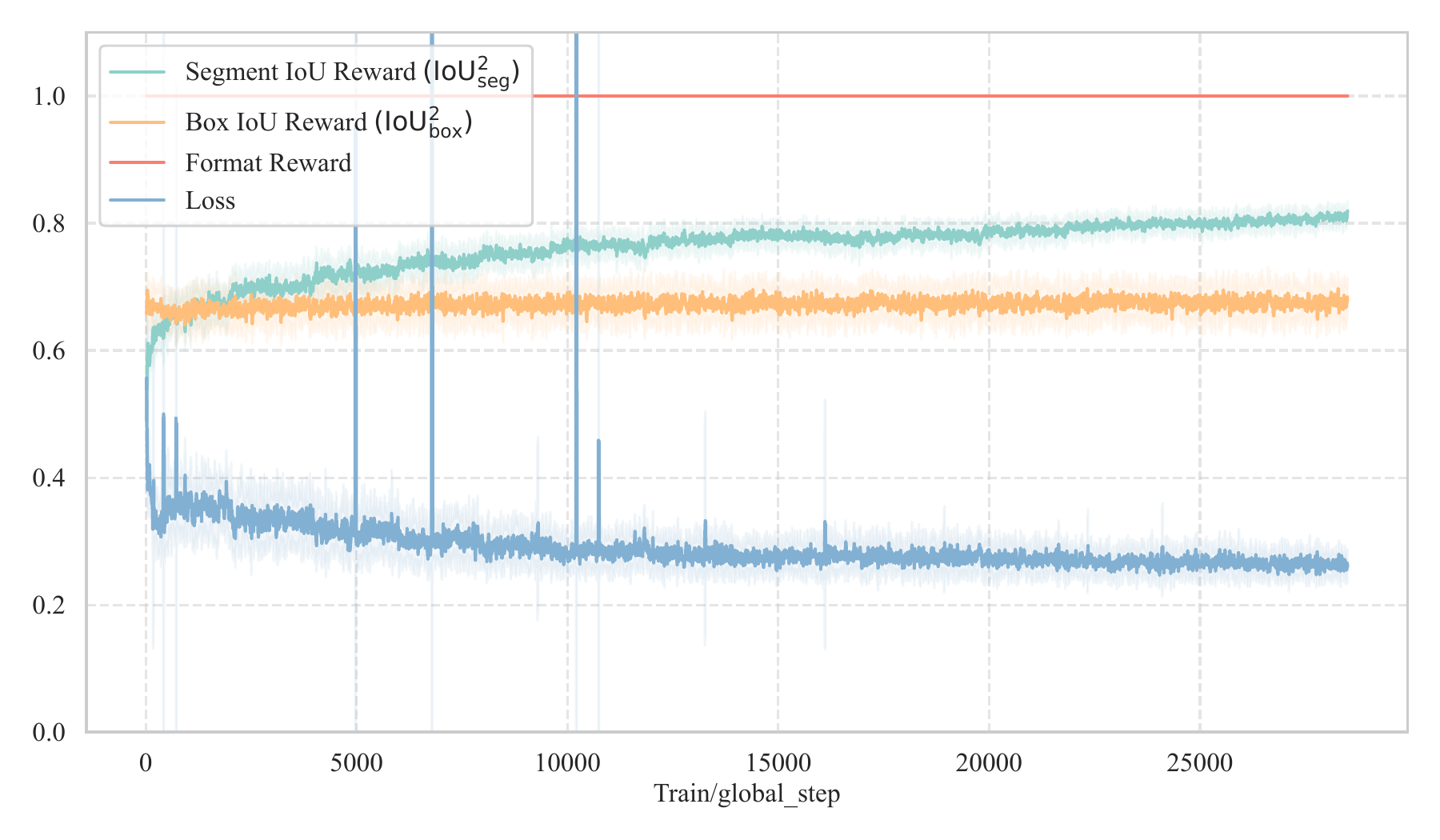}
    \caption{Over Unified GRPO training, the format reward, segment reward, box reward, and the loss curves indicate steady reward gains and consistent loss decreases.}
    \label{fig:reward_loss_curve}
\end{figure*}

The GRPO training process demonstrates stable and effective policy learning, as evidenced by the concurrent trends in reward and loss (Fig.~\ref{fig:reward_loss_curve}). We define the IoU reward as the squared segment IoU and the squared box IoU, i.e., reward is $\text{IoU}_\text{seg}^2$ or $\text{IoU}_\text{box}^2$. We observe a steady increase in both format reward, segment IoU, and box IoU rewards, which correlates with the generation of better-formatted responses, higher-quality masks and more accurate grounding predictions. The reward improvements corresponds with a steady decrease in training loss, confirming the convergence of the optimization process.

\subsection{More Module Details}
Our multimodal large language model (MLLM) is Qwen2.5-VL-3B, whose embedding dimension is 2048. 
The context query module consists of 64 learnable embeddings, which serve as task-specific query slots for extracting and structuring multimodal reasoning signals from the MLLM.
The connector module is implemented as a 4-layer Qwen2.5-style Transformer with a hidden dimension of 2048, ensuring dimensional alignment with the MLLM outputs. 
It maps textual and segmentation prompts into a unified embedding space, enabling seamless interaction between the MLLM’s high-level reasoning and the SAM2 segmentation head.

As summarized in Tab.~\ref{tab:module_details}, the context query and connector together contribute only 4.9\% of the total parameters and 4.5\% of the overall inference latency. Despite being lightweight, these modules provide an effective and efficient bridge between multimodal understanding and mask generation, significantly enhancing grounding and segmentation performance with minimal computational overhead.

\subsection{More Visualization Examples}

In this section, we provide additional qualitative examples to demonstrate the capabilities and behavior of our model under various challenging scenarios.
We present visualizations covering referring expression segmentation, reasoning-based segmentation, and cases with imperfect or misleading input conditions.

\subsubsection{Spatial Understanding}

\begin{figure*}[ht]
    \centering
    \includegraphics[width=.95\textwidth]{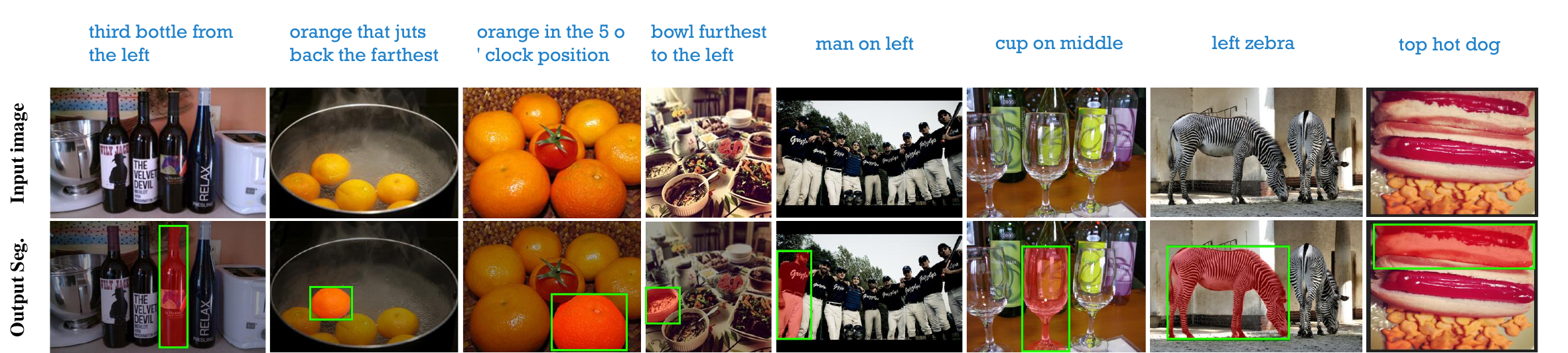}
    \caption{Example of spatial understanding. LENS correctly segments the target object based on spatial cues in the referring expression.}
    \label{fig:spatial_understanding}
\end{figure*}

As shown in Fig.~\ref{fig:spatial_understanding}, we showcase examples where the model successfully interprets spatial cues in the referring expressions, such as ``third bottle from the left" or ``the orange in the 5 o'clock position". 
These cases highlight the model's ability to integrate language-based spatial relations with visual layout.

\subsubsection{Multi-object Disambiguation}

\begin{figure*}[ht]
    \centering
    \includegraphics[width=.95\textwidth]{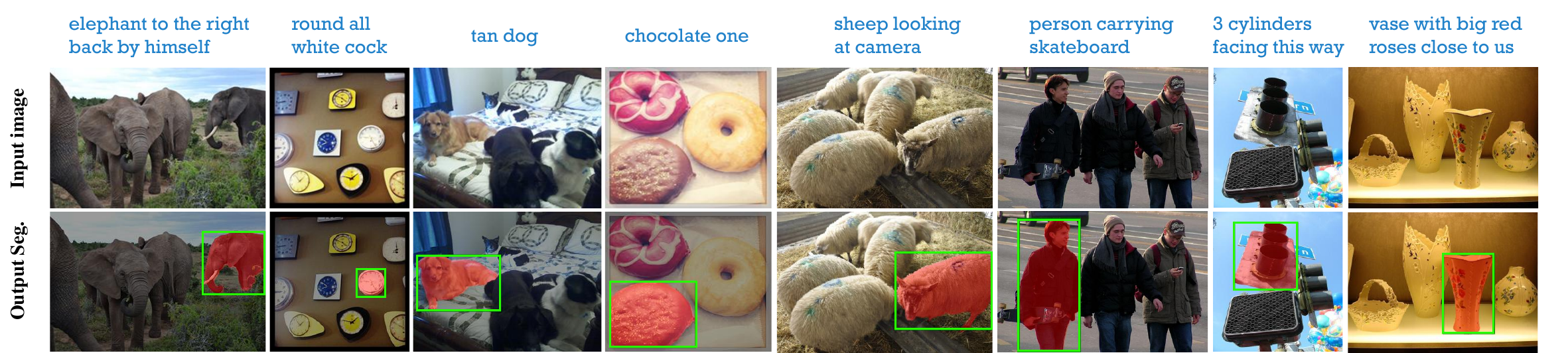}
    \caption{Example of disambiguation among multiple similar objects. 
    LENS recognizes distinguishing attributes in the referring expression and accurately identifies the intended target.}
    \label{fig:multi_object_disambiguation}
\end{figure*}

As shown in Fig~\ref{fig:multi_object_disambiguation}, we present examples in which multiple similar objects are present in the scene, and the referring expression must disambiguate between them. 
Our method is able to correctly segment the intended target by leveraging distinguishing attributes in the language input.

\subsubsection{Complex Referring Expressions}

\begin{figure*}[ht]
    \centering
    \includegraphics[width=.95\textwidth]{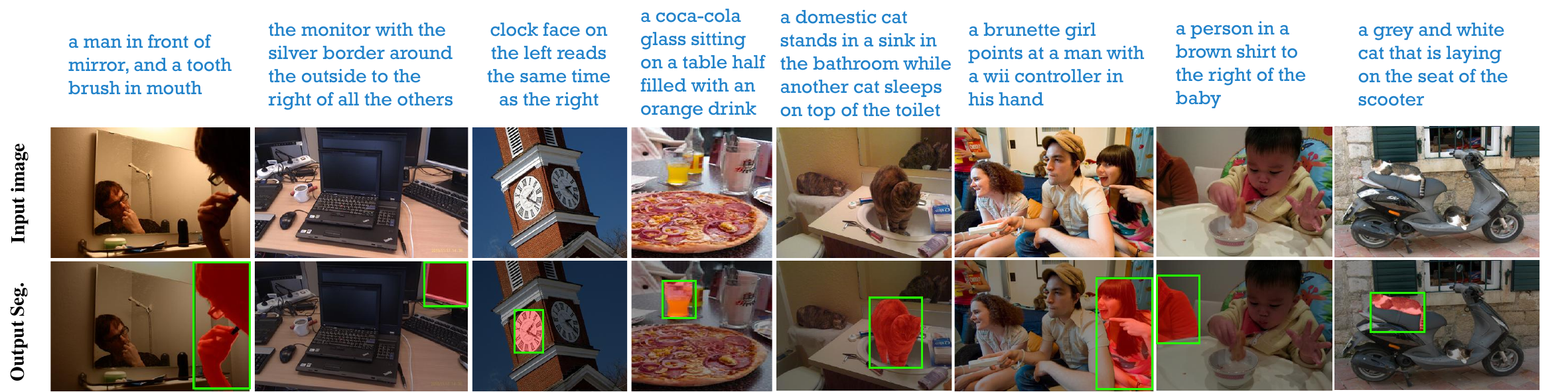}
    \caption{Example of handling complex referring expressions. LENS can process multiple attributes or relationships in the referring.}
    \label{fig:complex_referring}
\end{figure*}

As shown in Fig.~\ref{fig:complex_referring}, we provide examples involving long, compositional, and descriptive expressions that require the model to understand multiple attributes and relations. 
These cases demonstrate the model’s ability to handle syntactic complexity and semantic richness.

\begin{figure*}[t]
    \centering
    \includegraphics[width=.95\textwidth]{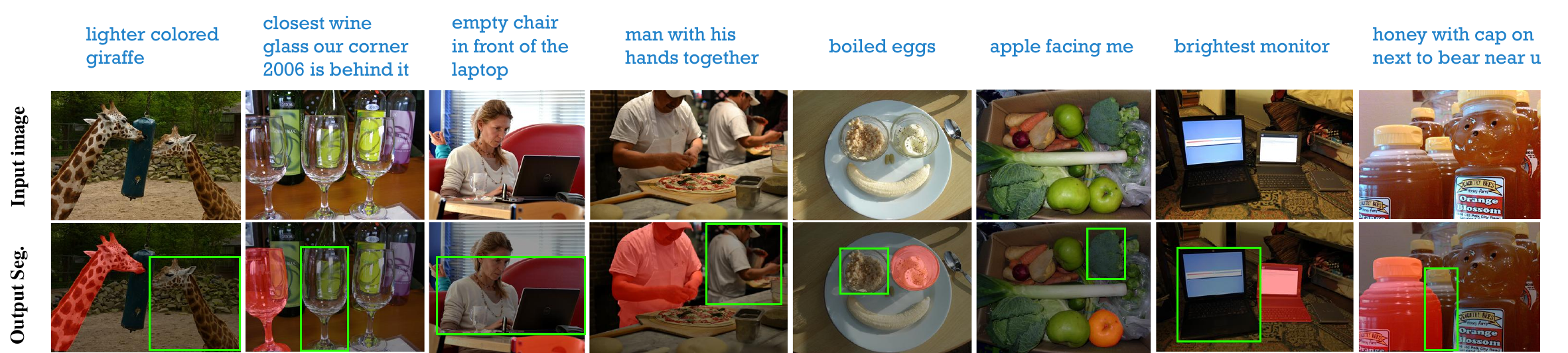}
    \caption{Example of segmentation under misguided context. When the initial context box is incorrect or ambiguous, LENS can leverage multi-query context embeddings to generate an accurate segmentation mask, demonstrating robustness to context errors.}
    \label{fig:misguided_context}
\end{figure*}

\paragraph{Segmentation under Misguided Context}
As shown in Fig.~\ref{fig:misguided_context}, we highlight cases where the context box is incorrect or ambiguous. 
Despite the error, the segmentation module is able to generate accurate masks by leveraging rich multi-query context embeddings. 
These examples showcase the robustness of our unified framework to noisy or part of misleading contexts.

\subsubsection{Coarse Annotations}

\begin{figure*}[t]
    \centering
    \includegraphics[width=.95\textwidth]{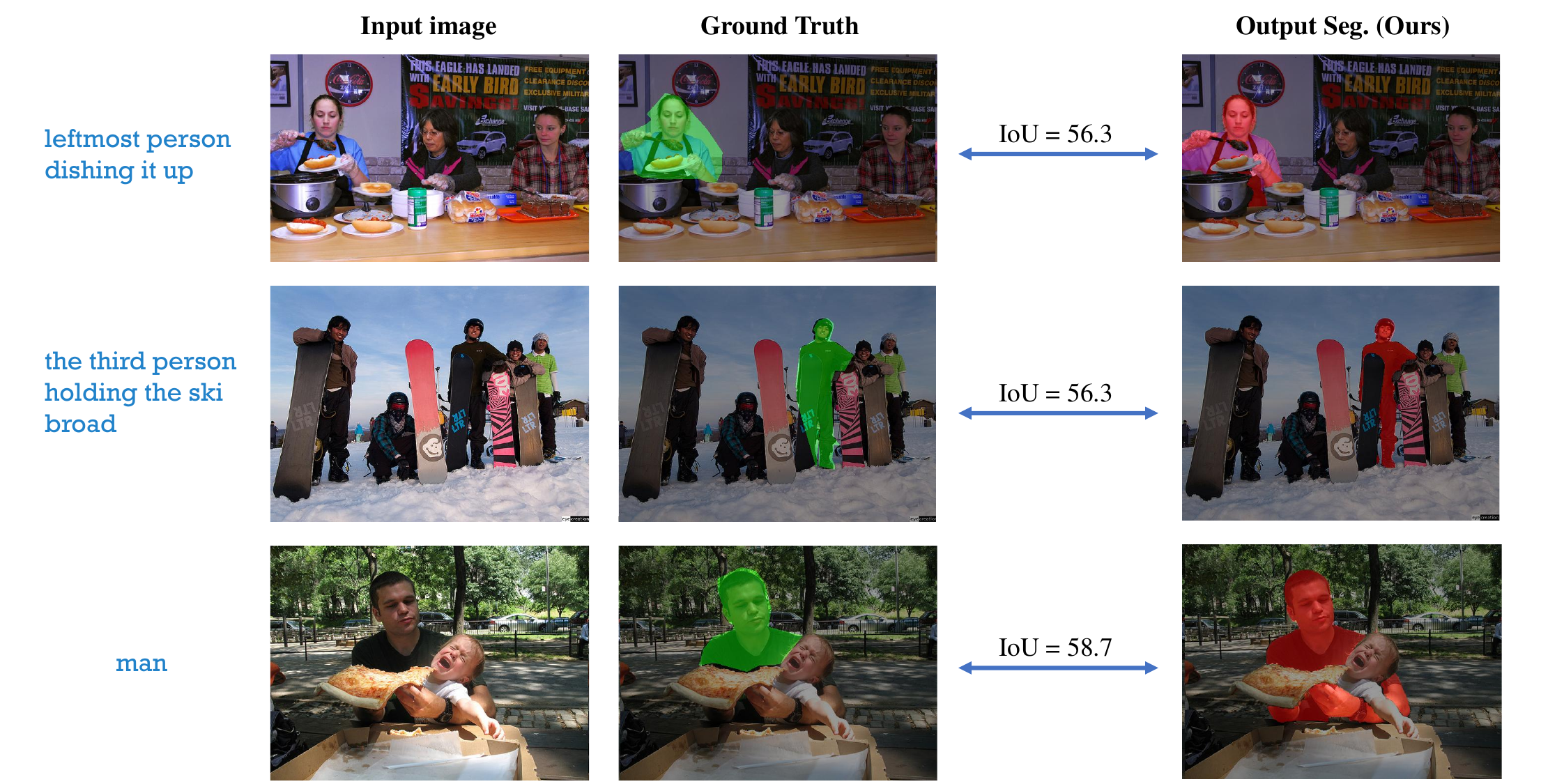}
    \caption{Example of coarse annotations. LENS predicts a mask that captures fine-grained object boundaries and surpasses the coarse or imprecise ground-truth mask annotations.}
    \label{fig:coarse_annotation}
\end{figure*}

As shown in Fig.~\ref{fig:coarse_annotation}, sometimes ground-truth annotations are coarse, but the proposed method can predict fine-grained and precise masks. 
These examples illustrate the model's ability to produce detailed masks that may better reflect the true object boundaries, despite noisy or imprecise supervision.

\subsubsection{Reasoning Segmentation}
\begin{figure*}[t]
    \centering
    \includegraphics[width=.95\textwidth]{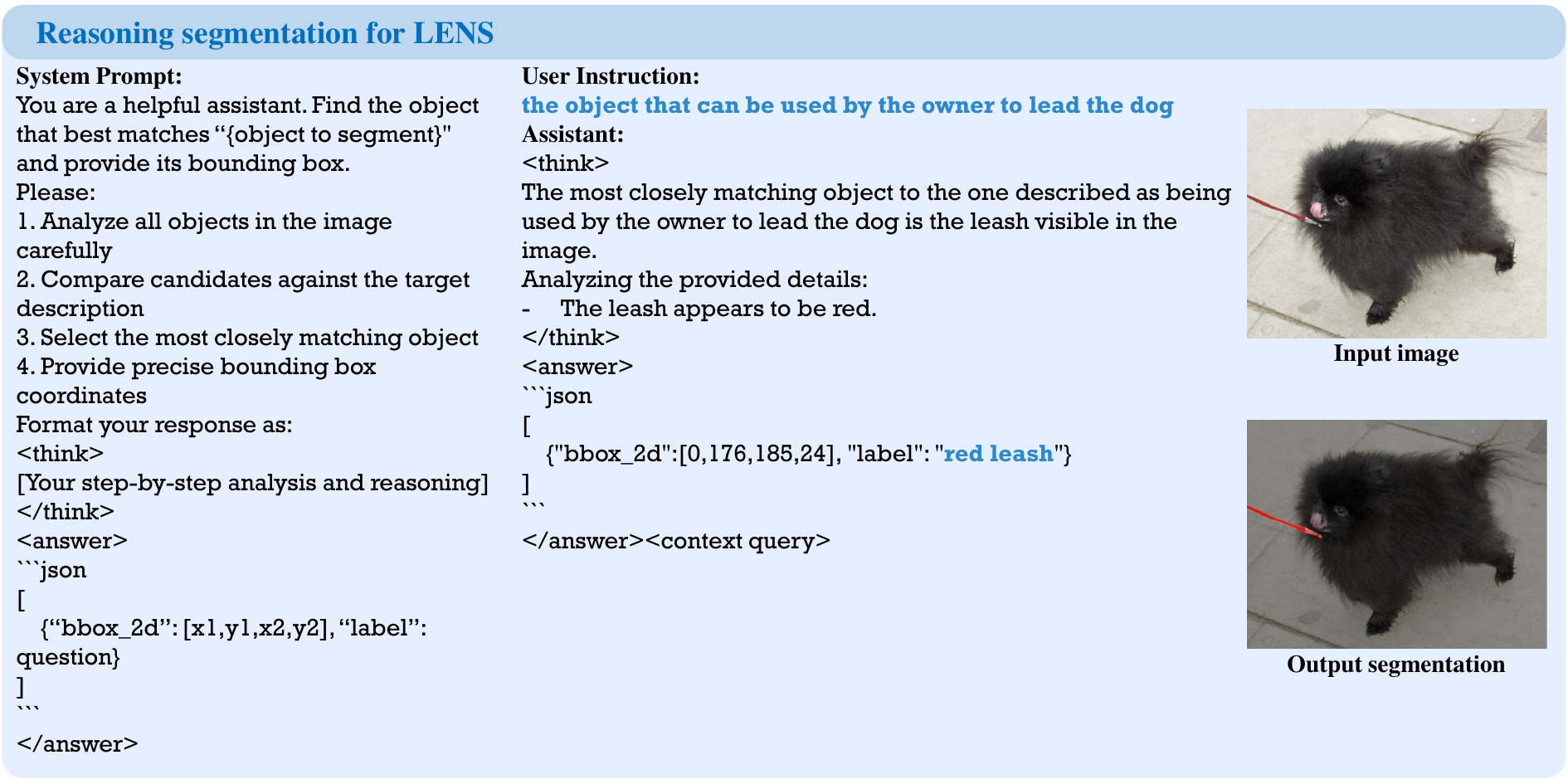}
    \caption{Example of reasoning segmentation. LENS successfully infers and segments the target based on a complex instruction.}
    \label{fig:reasoning_segmentation}
\end{figure*}

As shown in Fig.~\ref{fig:reasoning_segmentation}, we further visualize samples that require reasoning beyond simple visual matching, such as temporal states (e.g., ``the person who is eating") or implicit object relationships. 
The results demonstrate the model’s capacity for complex reasoning segmentation.

\subsection{More Discussion about Failure Cases}
\begin{figure*}[ht]
    \centering
    \includegraphics[width=.81\textwidth]{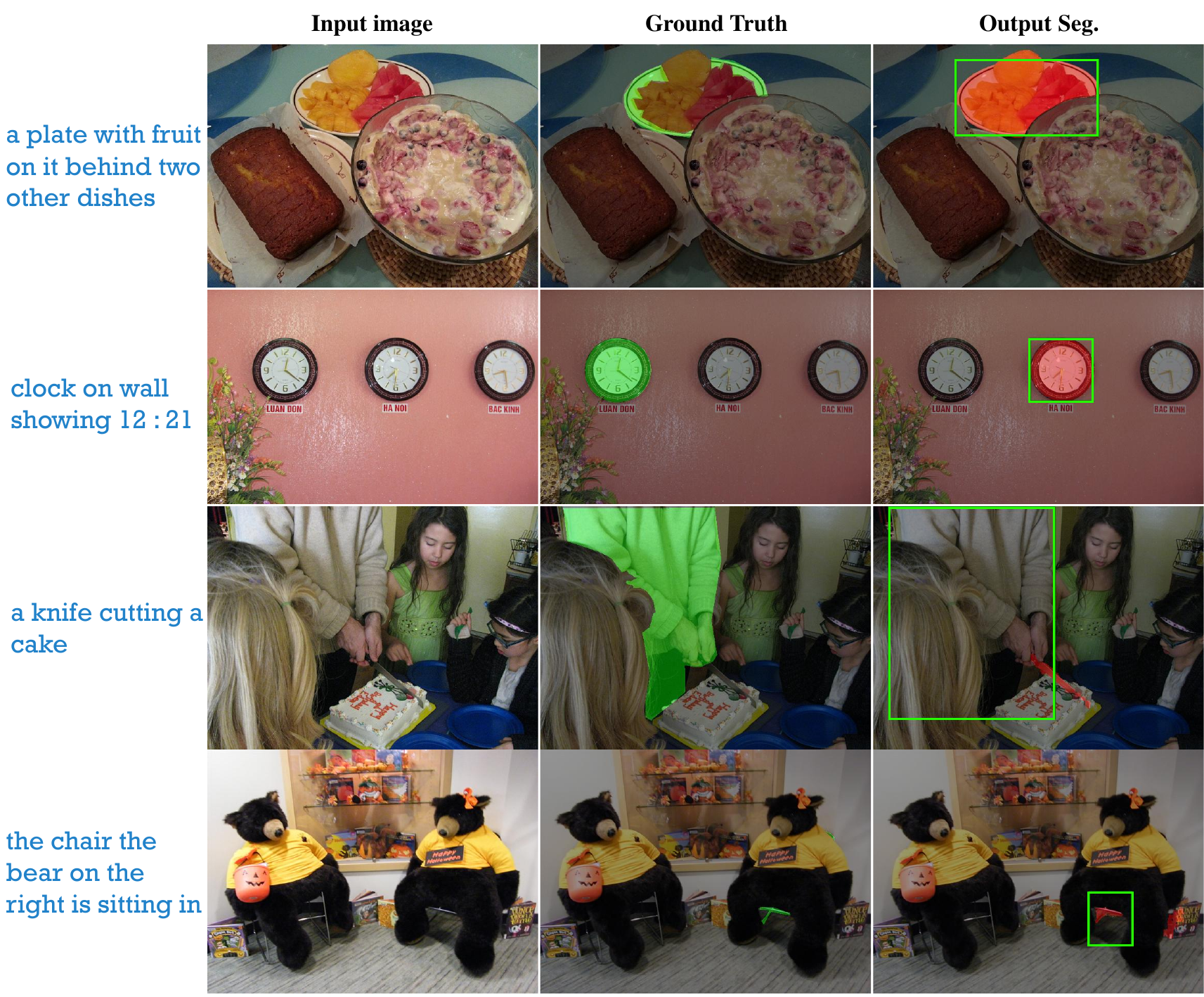}
    \caption{Failure cases: (1) granularity confusion, (2) wrong-instance selection, 
(3) annotation error, and (4) small object difficulty.}
    \label{fig:failure_cases}
\end{figure*}

Our method may fail in cases where the instruction requires fine-grained reasoning, 
refers to ambiguous attributes, or when the target object is very small. Fig.~\ref{fig:failure_cases} summarizes several representative failure modes. (1) Granularity confusion: when the instruction refers to “a plate with fruit on it,” our model segments both the plate and the fruit, whereas the ground truth only annotates the plate. This arises from the lack of part-level supervision. (2) Wrong-instance selection: for the instruction “the clock showing 12:21,” the model selects a different clock instance with mismatched attributes. (3) Annotation error: in “a knife cutting a cake,” the ground truth incorrectly labels the person holding the knife instead of the knife itself. (4) Small object difficulty: for “the chair the bear on the right is sitting in,” 
the target chair is very small in the image, which makes precise segmentation challenging and can lead to over-segmentation.

\end{document}